\documentclass[runningheads]{llncs}
\usepackage{graphicx}
\graphicspath{{Figures/}}
\usepackage[colorlinks,citecolor=green,linkcolor=red]{hyperref}
\usepackage{amsmath}
\usepackage{amsfonts}
\usepackage{booktabs}
\usepackage[misc,geometry]{ifsym}

\usepackage{colortbl}
\usepackage{adjustbox}

\begin{document}
\title{DomainSiam: Domain-Aware  Siamese Network  for Visual Object Tracking}

\author{Mohamed H. Abdelpakey   $^{(\textrm{\Letter})}$ \and
Mohamed S. Shehata  
}

\authorrunning{M. H. Abdelpakey et al.}
\institute{Memorial University of Newfoundland,\\ St. John's, NL A1B 3X5, Canada\\
\email{mha241@mun.ca}\\
}

\maketitle              %
\begin{abstract}

Visual object tracking is a fundamental task in the field of computer vision. Recently, Siamese trackers have achieved \textit{state-of-the-art} performance on recent benchmarks. However, Siamese trackers do not fully utilize semantic and objectness information from pre-trained networks that have been trained on the image classification task. Furthermore, the pre-trained Siamese architecture is sparsely activated by the category label which leads to unnecessary calculations and overfitting. In this paper, we propose to learn a Domain-Aware, that is fully utilizing semantic and objectness information while producing a class-agnostic using a ridge regression network.  Moreover, to reduce the sparsity problem, we solve the ridge regression problem with a  differentiable weighted-dynamic loss function. Our tracker, dubbed \textit{DomainSiam}, improves the feature learning in the training phase  and generalization capability  to other domains.
Extensive experiments are performed on five tracking benchmarks including OTB2013 and OTB2015 for a validation set; as well as the  VOT2017, VOT2018, LaSOT, TrackingNet,  and GOT10k for a testing set. \textit{DomainSiam} achieves a \textit{state-of-the-art} performance on these benchmarks while running at 53 FPS.

\end{abstract}

\section{Introduction}

Tracking is a fundamental  task in many computer vision tasks such as surveillance \cite{kendall2015posenet}, computer interactions \cite{molchanov2016online} and image understanding \cite{lenc15understanding}. The objective of tracking is to find the trajectory of the object of interest over time. This is a challenge since the object of interest undergoes appearance changes such as occlusions, motion blur, and background cluttering \cite{wu2015object,alahari2015thermal}. Recent deep trackers such as CFNet \cite{valmadre2017end} and DeepSRDCF \cite{danelljan2015convolutional} use  pre-trained networks which have been trained on image classification or object recognition.

In recent years, convolutional neural networks (CNNs) have achieved superior performance against hand-crafted trackers (e.g., CACF \cite{mueller2017context}, SRDCF \cite {danelljan2015learning}, KCF \cite{henriques2015high} and SAMF \cite{li2014scale}). Siamese trackers such as SiamFC \cite{bertinetto2016fully}, CFNet \cite{valmadre2017end}, SiamRPN \cite{li2018high}, and  DensSiam \cite{mohamed2018denssiam} learn   a similarity function to separate the foreground from its background. However, Siamese trackers do not fully utilize the semantic and objectness information from pre-trained networks that have been trained on image classification. In image classification, the class categories of the objects are pre-defined while in object tracking task the tracker needs to be class-agnostic while benefiting from semantic and objectness information. Moreover, the image classification increases the inter-class differences  while forcing the features to be insensitive to intra-class changes \cite{li2019target}.

In this paper, we propose DomainSiam to learn Domain-Aware, that is fully utilizing semantic and objectness information from a pre-trained network. DomainSiam consists of the DensSiam with a self-attention module \cite{mohamed2018denssiam} as a backbone network  and a regression network to select the most discriminative convolutuinal filters to leverage the semantic and objectness information. Moreover, we develop a differentiable  weighted-dynamic domain loss function to train the  regression network. The developed  loss function is monotonic, dynamic, and smooth with respect to its hyper-parameter, which can be reduced to \textit{$l_1$ or $l_2$} during the training phase. On the other hand, the shrinkage loss function \cite{lu2018deep} is static and it can not be adapted during the training phase.  Most regression networks solve the regression problem with static loss such as the closed-form solution if the input to the network is not high-dimensional or minimizing \textit{$l_2$}. The results  will be made available  \footnote{\url{https://vip-mun.github.io/DomainSiam}}.\\
To summarize, the main contributions of this paper are three-fold.    
\begin{itemize}
   \item[$\bullet$] A novel architecture is proposed for object tracking to capture the Domain-Aware features with semantic and objectness information. The proposed architecture enables the features to be robust to appearance changes. Moreover, it decreases the sparsity problem as it produces the most important feature space. Consequently, it decreases  the overhead calculations.\\  
   
   \item[$\bullet$] Developing a differentiable weighted-dynamic domain loss function specifically for visual object tracking to train the regression network to extract the domain channels that is activated by target category. The developed loss is monotonic with respect to its hyper-parameter, this will be useful in case of high dimensional data and non-convexity. Consequently, this will increase the performance of the tracker. 
   \item[$\bullet$] The proposed architecture tackles  the generalization capability from one domain to another domain (e.g., from ImageNet to VOT datasets).  
   \end{itemize} 
The rest of the paper is organized as follows. Related work is presented in  section \ref{Related work}. Section \ref{Proposed} details the proposed approach. We present the experimental results in Section \ref{Experiments}. Finally, section \ref{Conclusions } concludes the paper.

\section{Related work}
\label{Related work}

Recently, Siamese-based trackers have received significant attention especially in realtime tracking. In this section, we firstly introduce the \textit{state-of-the-art} Siamese-based trackers. Then, we briefly introduce the gradient-based localization guidance.\\\\
\textbf{Siamese-based Trackers}\\\\
The first Siamese network was first proposed in \cite{bromley1994signature} for signature verification. In general, Siamese network consists of two branches the target branch and the search branch; both branches share the same parameters. The score map which indicates the position of the object of interest is generated by the last cross-correlation layer. The pioneering work of using Siamese in object tracking is SiamFC \cite{bertinetto2016fully}. SiamFC searches the target image in the search image. Siamese Instance Search \cite{tao2016siamese} proposed SINT, which has the query  branch and the search branch, the backbone of this architecture is inherited from AlexNet \cite{krizhevsky2012imagenet}. CFNet \cite{valmadre2017end} improved SiamFC by adding a correlation layer to the target branch. SA-Siam \cite{he2018twofold} proposed two Siamese networks, the first network encodes the semantic information   and the second network encodes the appearance model, which is different from our architecture that has only one Siamese network. SiamRPN \cite{Li_2018_CVPR} formulated the tracking problem as a local one-shot detection. SiamRPN consists of a Siamese network as a feature extractor and a region proposal network which includes the classification branch and regression branch. DensSiam \cite{mohamed2018denssiam} used the Densely-Siamese architecture to make the Siamese network deeper while maintaining the performance of the network. DensSiam allows the low-level and high-level features to flow  within layers without vanishing gradients problem. Moreover, a self-attention mechanism was integrated to force the network to capture the non-local features. SiamMask \cite{wang2019fast} used Siamese networks for object tracking using augmentation loss to produce a binary segmentation mask. In addition, the binary segmentation mask locates the  object  of interest accurately. ATOM \cite{danelljan2019atom} proposed the Siamese network with  explicit components for target estimation and classification. The component is trained offline to maximize the overlapping between the estimated bounding box and the target. Most Siamese trackers do not fully utilize the semantic and ojectness information from pre-trained networks. \\\\
\textbf{Gradient-based Localization Guidance}\\\\
In this category of learning, the objective is to determine the most important channel of the network with respect to the object category. In an object classification task, each category  activates a set of certain channels. Grad-CAM \cite{selvaraju2017grad} used a gradient of any target logit (e.g., "cat") and using this gradient, determined the active category channel for this logit. The work in \cite{Zhou_2016_CVPR} demonstrated that the global average pooling of the gradients is implicitly acting as  Attention for the network; consequently, it can  locate the object of interest accurately. 
\section{Proposed Approach}
\label{Proposed}

\begin{figure}
\centering
\includegraphics[width=0.9\columnwidth]{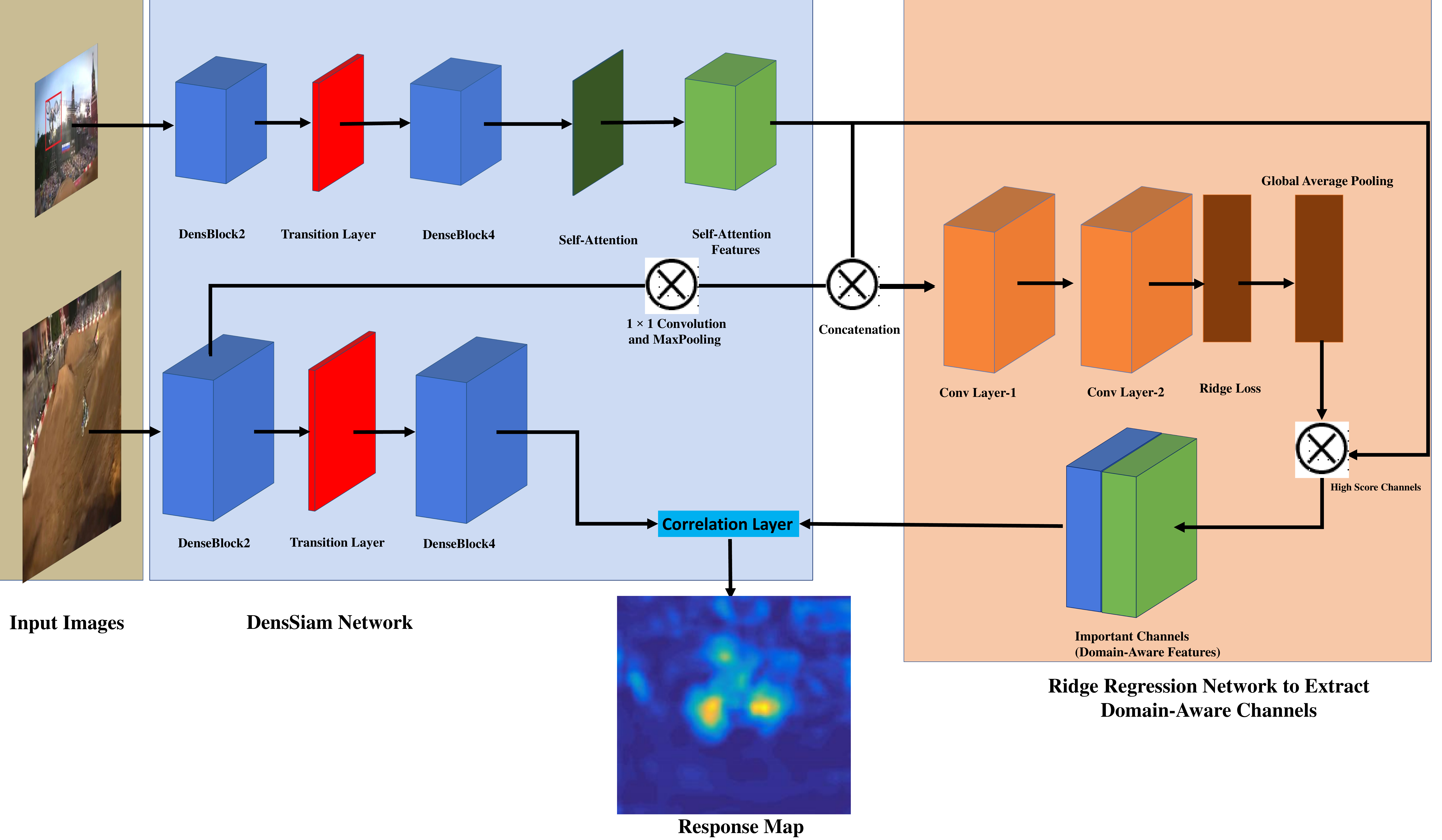}

\caption{The architecture of DomainSiam tracker. It consists of three blocks, the input images block which includes the target image and search image, DensSiam network with a Self-Attention module at the end of the target branch, and the Ridge Regression Network that highlights the important channels  and produces the Domain-Aware  features. The response map is produced by the  correlation layer which is the final layer. The correlation layer calculates the correlation between the Domain-Aware channels and search branch features which is represented by DenseBlock4.}
\label{DomainSiam}
\end{figure}

We propose DomainSiam for visual object tracking, the complete pipeline is shown in Fig. \ref{DomainSiam}. The DensSiam with the Self-Attention network is used as a feature extractor, however, in any Siamese network these features do not fully utilize the semantic and objectness information. Furthermore, the channels in Siamese networks are sparsely activated. We use the ridge regression network with a differentiable weighted-dynamic loss function to overcome the previous problems.

\begin{figure}
\centering
\includegraphics[width=0.7\columnwidth]{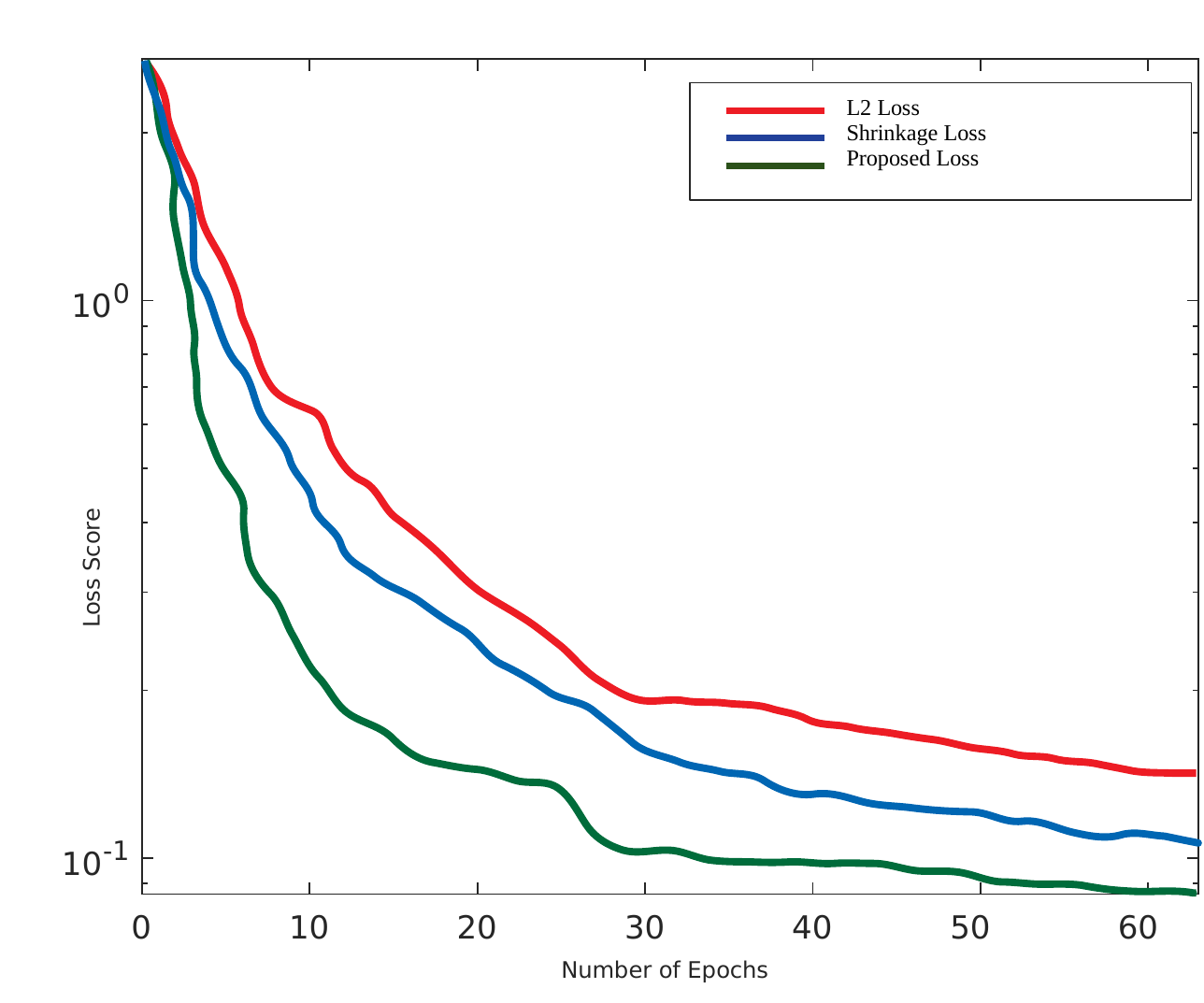}

\caption{A comparison of convergence speed on $L_2$ loss, Shrinkage loss \cite{lu2018deep}, and our proposed loss function. The average loss is calculated on a batch of eight samples on VOT2018 \cite{kristan2018sixth} dataset.}
\label{Losscurve}
\end{figure}

\subsection{Ridge Regression Network with Domain-Aware Features}

In Fig. \ref{DomainSiam}, the pipeline is divided into three blocks, the input block to the target branch and the search branch, the DensSiam block which has the same architecture in \cite{mohamed2018denssiam}, and the ridge regression network. The DensSiam network produces two feature maps for target and search images, respectively. Imbalanced distribution of the training data makes the feature maps produced by Siamese networks less discriminative as there is a high number of easy samples compared to the hard samples. Siamese networks use pre-trained networks which have been trained on other tasks (e.g., classification and recognition). These networks increase inter-class differences and is also insensitive to intra-class variations. Consequently, this property decreases the performance of Siamese networks as the tracker needs to be more discriminative to the same object category. Moreover, the pre-trained network is sparsely activated by the object category. In other words, in the feature channels/maps there are only a few active channels that correspond to the object category. The regression network in Fig. \ref{DomainSiam} highlights the importance of each channel in the feature map to the object of interest and discards the others.\\
In Fig. \ref{DomainSiam}, the ridge  regression network regresses all samples in the input image patch to their soft labels by optimizing the following objective function. 
\begin{equation} \label{loss_reg}
\underset{w}{\arg \min }\left\|W * X_{i, j} - Y(i, j)\right\|^{2}+\lambda\|W\|^{2}
\end{equation} 
Where $\|*\|$ denotes the convolution operation, $W$ is the weight of the regression network, $\mathbf{X} \in \mathbb{R}^{N \times D}$   is the input features and   $\mathbf{Y} \in \mathbb{R}^{N \times D}$    is the soft label. Gaussian distribution is used as a soft label map and its centre is aligned to  the target center and $\lambda > 0$ is the regularization parameter.  
\begin{equation} \label{gaussian}
Y(i, j)=e^{-\frac{i^{2}+j^{2}}{2 \sigma^{2}}}
\end{equation}
Where $(i,j)$ is the location corresponding to the target location and $\sigma$ is the Gaussian kernel width. The closed-form analytic solution for equation \ref{loss_reg} is defined as

\begin{equation} \label{closed_solution}
W=\left({X}^{\top} {X}+\lambda {I}\right)^{-1} {X}^{\top} {Y}
\end{equation}
The optimal solution of $W$ can be achieved by equation \ref{closed_solution}; however, solving this equation is computationally expensive as ${X}^{\top} {X} \in \mathbb{R}^{D \times D}$. Instead, we use the ridge regression network with the proposed loss function to solve equation \ref{loss_reg}.

\subsection{Ridge Regression Optimization}

In Fig. \ref{DomainSiam}, the ridge regression network consists of two convolutional layers, ridge loss and the global average pooling. The global average pooling encourages the proposed loss function to localize the object of interest accurately compared to the global max pooling. It is worth mentioning that both global average pooling and global max pooling have similar performances on object classification tasks. As shown in Fig. \ref{DomainSiam}, in the last block, the Domain-Aware feature space  is calculated by 

\begin{equation} \label{importance}
\delta_i = GAP ({\partial L}/{\partial F_{i}})
\end{equation}
Where $\delta$ is the Domain-Aware non-sparse features, $GAP$ is the global  average pooling, $L$ is the domain-dynamic loss function which will be discussed later, and $F$ is the input feature channel of the $i^{th}$ channel to the ridge regression network. Let the objective function of the ridge regression network be $x$
\begin{equation} \label{x_loss_reg}
x= \|W * X_{i, j} - Y(i, j)\|^{2}+\lambda\|W\|^{2}
\end{equation}
 We propose a  differentiable weighted-dynamic loss function for visual object tracking to solve equation \ref{x_loss_reg}, inspired by \cite{barron2019general} that uses  a general loss function for variational autoencoder, monocular depth estimation, and global registration, as follows.

\begin{equation} \label{new_loss}
L(x, \alpha)= \frac{|\alpha-2| }{\alpha}e^{ay}\left(\left(\frac{x^{2}}{|\alpha-2|}+1\right)^{\alpha / 2}-1\right)
\end{equation}
where $a \in [0,1]$ is a hyper-parameter, $y$ is the regression target of a sample, and $\alpha \in \mathbb{R}$ is the parameter that controls the robustness of the loss. The exponent term in this loss function tackles the imbalanced distribution of the training set by assigning a higher weight to hard samples. The imbalanced data occurs when the number of easy samples (background) is extremely higher than the hard samples (foreground).\\
The advantage of this loss function over  equation  \ref{loss_reg} and  equation \ref{closed_solution} is it can  automatically adjust the robustness during the training phase. This advantage comes from the $\alpha$ parameter. For example, at $\alpha = 2$ the equation  \ref{new_loss} becomes $L_2$
\begin{equation}
\lim _{\alpha \rightarrow 2} L(x, \alpha)=\frac{e^{ay}}{2}  x ^{2}
\end{equation} 
Similarly, when $\alpha = 1$, the equation \ref{new_loss} becomes $L_1$

\begin{equation}
L(x, \alpha)=  (\sqrt{x ^{2}+1})e^{ay} -1
\end{equation}
Another advantage of equation \ref{new_loss} is becoming Lorentzia loss function \cite{black1996robust}  by allowing $\alpha= 0$  as  follows  

\begin{equation}
\lim _{x \rightarrow 0} L(x, \alpha)=\log \left(\frac{1}{2}x^{2}+1\right)e^{ay}
\end{equation}
As noticed before that the proposed loss function is dynamic, this allows the network   to also  learn  a robust representation. The gradient of the equation \ref{new_loss}  with respect to  $\alpha$ is always positive. Consequently, this property makes the loss monotonic with respect to $\alpha$ and useful for non-convex optimization.
\begin{equation}
\frac{\partial L}{\partial \alpha}(x, \alpha) \geq 0
\end{equation}
The final proposed loss function is given by
\begin{equation}\label{final_loss}
L(x, \alpha)=\left\{\begin{array}{ll}{\frac{e^{ay}}{2} x^{2}} & {\text { if } \alpha=2} \\ {\log \left(\frac{1}{2}(x )^{2}+1\right) e^{ay}} & {\text { if } \alpha=0} \\ {(1-\exp \left(-\frac{1}{2}(x)^{2}\right))e^{ay}} & {\text { if } \alpha=-\infty} \\ {\frac{|\alpha-2|}{\alpha} e^{ay}\left(\left(\frac{(x )^{2}}{|\alpha-2|}+1\right)^{\alpha / 2}-1\right)} & {\text { otherwise }}\end{array}\right.
\end{equation}
Fig. \ref{Losscurve} shows that the optimization over the proposed loss function achieves faster convergence speed, while in Shrinkage loss function that is proposed in \cite{lu2018deep} and the original ridge regression loss function \ref{loss_reg} ($l_2$), the convergence speed is slower. 
The importance of each channel in the feature map is calculated by plugging equation \ref{final_loss} into equation \ref{importance}.
It is worth mentioning that the output feature map of the ridge regression network contains only the activated channels that have the most semantic and objectnes information corresponding to the object category. The Domain-Aware features and the feature channels from denseBlock4 are fed into the correlation layer to calculate the similarity and produce the response map. 

\section{Experimental Results}
\label{Experiments}
The benchmarks are divided into two categories, the validation set including OTB2013 \cite{wu2013online} and OTB2015 \cite{wu2015object}, and the testing set including   VOT2017 \cite{kristan2017visual}, VOT2018 \cite{kristan2018sixth}, and  GOT10k \cite{huang2018got}. We introduce the implementation details in the next sub-section and then we compare the proposed tracker to the \textit{state-of-the-art} trackers.
\subsection{Implementation Details}
We used the pre-trained DensSiam network (DenseBlock2 and DenseBlock4) that has been trained  on Large Scale Visual Recognition Challenge (ILSVRC15) \cite{russakovsky2015imagenet}. ILSVRC15 has over 4000 sequences with approximately 1.3 million frames with their labels. The DomainSiam, which has been trained on 1000 classes, can  benefit from this  class  diversity. We implemented DomainSiam in Python using PyTorch framework \cite{paszke2017pytorch}. Experiments are performed on  Linux with  a Xeon E5 $@$2.20 GHz CPU and a Titan XP GPU. Testing speed of DomainSiam is 53 FPS which is beyond realtime speed.\\
\textbf{Training}. The ridge regression network is trained with its proposed loss function separately from the Siamese network  with 70 epochs. The highest scores associated with  100 channels are selected as the Domain-Aware features. The training is applied with a momentum of 0.9, the batch size  of 8 images, and the learning rate is annealed geometrically at each epoch from $10^{-3}$ to $10^{-8}$.\\
\textbf{Tracking Settings}. The initial scale variation is $O^{s}$ where $O=1.0375$ and $s=\{-2,0,2\}$. We adopt the target image size of $127 \times 127$ and the search image size of $255 \times 255$ with a linear interpolation to update the scale with a factor of $0.435$.

\begin{table*}[!t]
\centering
\caption{Comparison with the state-of-the-art trackers on VOT2017 in terms of Accuracy (A), expected Average Overlap (EAO), and Robustness (R).}\label{VOT2017}
\begin{tabular}{l|c|c|c|r}
\specialrule{1.2pt}{0pt}{0pt}
Tracker &  A$\uparrow$ & EAO$\uparrow$  &R $\downarrow$ &  FPS\\
\specialrule{1.2pt}{0pt}{0pt}
CSRDCF++&0.453&0.229&0.370& $>$ 25\\
SAPKLTF& 0.482&0.184&0.581& $>$ 25 \\
Staple & 0.530&0.169& 0.688&  $>$ 80 \\
ASMS & 0.494& 0.169&0.623 &  $>$ 25\\
SiamFC & 0.502&0.188& 0.585& 86\\

SiamDCF & 0.500 &0.473& 0.249&60\\
ECOhc& 0.494&0.435&0.238&60\\
DensSiam& 0.540& 0.350& 0.250 &60 \\ 
\hline
\textbf{DomainSiam(proposed)}  & \textbf{0.562}& \textbf{0.374} &\textbf{0.201}& 53 \\ 

\specialrule{1.2pt}{0pt}{0pt}
\end{tabular}
\end{table*}

\begin{table*}[!t]
\centering
\caption{Comparison with \textit{state-of-the-art} trackers on VOT2018 in terms of Accuracy (A), expected Average Overlap (EAO), and Robustness (R).}\label{VOT2018}
\begin{tabular}{l|c|c|c|r}
\specialrule{1.2pt}{0pt}{0pt}
Tracker &  A$\uparrow$ & EAO$\uparrow$ &  R$\downarrow$ & FPS\\

\hline
ASMS \cite{vojir2014robust}& 0.494&  0.169&0.623 &25\\
SiamRPN \cite{li2018high}&0.586&0.383& 0.276&160 \\
SA$\_$Siam$\_$R \cite{he2018twofold}&0.566 & 0.337&0.258&50 \\
FSAN \cite{kristan2018sixth} &0.554&0.256 &0.356&30\\
CSRDCF \cite{lukezic2017discriminative}&0.491&0.256&0.356&13\\
SiamFC \cite{bertinetto2016fully}& 0.503& 0.188& 0.585&86\\
SAPKLTF \cite{kristan2018sixth}& 0.488&0.171& 0.613&25 \\
DSiam \cite{guo2017learning}& 0.215 & 0.196&0.646&25 \\
ECO \cite{danelljan2017eco}& 0.484&0.280&0.276&60 \\

\hline

\textbf{DomainSiam(proposed)} &\textbf{0.593} & \textbf{0.396} &\textbf{0.221}&53 \\

\specialrule{1.2pt}{0pt}{0pt}
\end{tabular}
\end{table*}

\begin{table*}[!t]
	\centering
	\caption{Comparisons with \textit{state-of-the-art} trackers on TrackingNet dataset in terms of the Precision (PRE),  Normalized Precision (NPRE), and Success.}
	\label{table:TrackingNet}

	 \begin{tabular}{l|c|c|r}
\specialrule{1.2pt}{0pt}{0pt}
Tracker &  PRE $\uparrow$ & NPRE $\uparrow$ &  SUC.$\uparrow$ \\
		\specialrule{1.2pt}{0pt}{0pt}
		
		Staple\_CA~\cite{mueller2017context} & 0.468 & 0.605 & 0.529 \\
		BACF~\cite{galoogahi2017learning}  & 0.461 & 0.580  & 0.523 \\
		
		MDNet~\cite{nam2016learning} & 0.565 & 0.705 &0.606 \\
		CFNet~\cite{valmadre2017end} & 0.533 & 0.654 & 0.578 \\
		SiamFC~\cite{bertinetto2016fully} & 0.533 & 0.663 & 0.571 \\
		
		SAMF~\cite{li2014scale}  & 0.477 & 0.598 & 0.504 \\
		ECO-HC~\cite{danelljan2017eco} & 0.476 & 0.608 & 0.541 \\
		Staple~\cite{bertinetto2016staple} & 0.470  & 0.603 & 0.528 \\

		ECO~\cite{danelljan2017eco}   & 0.492 & 0.618 & 0.554 \\
		CSRDCF~\cite{lukezic2017discriminative} & 0.480  & 0.622 & 0.534 \\

		\hline
		\textbf{DomainSiam(proposed)}  &\textbf{0.585} & \textbf{0.712} & \textbf{0.635}  \\
		\specialrule{1.2pt}{0pt}{0pt}
	\end{tabular}%
    \end{table*}%

\begin{table*}[!t]
	\centering
	\caption{Comparison  with \textit{state-of-the-art} trackers on  LaSOt dataset in terms of the Normalized Precision and Success.}
	\label{table:lasot}

	 \begin{tabular}{l|c|r}
\specialrule{1.2pt}{0pt}{0pt}
Tracker & Norm.\ Prec. (\%)$\uparrow$& Success (\%)$\uparrow$  \\
		\specialrule{1.2pt}{0pt}{0pt}
		MDNet \cite{nam2016learning}& 46.0& 39.7\\
DaSiam \cite{zhu2018distractor}& 49.6& 41.5\\
STRCF \cite{li2018learning}& 34.0&30.8 \\
SINT  \cite{tao2016siamese}& 35.4& 31.4\\
		StrucSiam \cite{zhang2018structured} &41.8 &33.5 \\
SiamFC \cite{bertinetto2016fully} &42.0 & 33.6\\
VITAL \cite{song2018vital}& 45.3& 39.0\\

ECO \cite{DanelljanCVPR2017}& 33.8& 32.4\\
DSiam \cite{guo2017learning}& 40.5& 33.3\\

\hline
\textbf{DomainSiam(proposed)}	&53.7 &43.6 \\

		\specialrule{1.2pt}{0pt}{0pt}
	\end{tabular}%
    \end{table*}%
      
\begin{table*}[!htb]
\centering
\caption{Comparison \textit{state-of-the-art} trackers on  GOT10k dataset  in terms of Average Overlap (AO), and Success Rates (SR) at overlap thresholds of 0.50 and 0.75.}
\label{table:GOT10k}

 \begin{adjustbox}{max width=\textwidth}
\begin{tabular}{|c|c|c|c|c|c|c |c |c|}
\specialrule{1.2pt}{0pt}{0pt}
TRACKER  & DomainSiam (proposed)	&      CFNet & SiamFC & GOTURN & CCOT  &  ECO  &  HCF  & MDNet \\ \hline
  AO     & \textbf{	0.414}					&      0.374 & 0.348  & 0.347  & 0.325 & 0.316 & 0.315 & 0.299 \\ \hline
SR(0.50) & 	\textbf{0.451}					&      0.404 & 0.353  & 0.375  & 0.328 & 0.309 & 0.297 & 0.303 \\ \hline
SR(0.75) & \textbf{0.214}			    	&      0.144 & 0.098  & 0.124  & 0.107 & 0.111 & 0.088 & 0.099 \\ \specialrule{1.2pt}{0pt}{0pt} 		
\end{tabular} 

   \end{adjustbox}
\end{table*}     
\subsection{Comparison with the State-of-the-Arts}

In this section we use five benchmarks to evaluate DomainSiam against \textit{state-of-the-art} trackers. We use  VOT2017 \cite{kristan2017visual},  VOT2018 \cite{kristan2018sixth},  LaSOT \cite{fan2019lasot},  TrackingNet \cite{muller2018trackingnet}, and GOT10k \cite{huang2018got}.\\ 
\textbf{Results on VOT2017 and VOT2018}\\ The results on the VOT dataset in Table \ref{VOT2017} and Table \ref{VOT2018} are given by VOT-Toolkit. DomainSiam   outperforms the \textit{state-of-the-art} trackers listed in both tables. It is worth mentioning that DoaminSiam is about $2\%$ higher than the DensSiam tracker in terms of   Accuracy (A) and Expected Average Overlap  (EAO)  in Table \ref{VOT2017} while running at 53 frames per second. Table \ref{VOT2018} shows that DomainSiam has a robustness of $0.221$ which is about $5\%$ higher than the second  best tracker (SiamRPN) while outperforming all other trackers in terms of accuracy and expected average overlap.\\
\textbf{Results on TrackingNet Dataset}\\
This is a large-scale  dataset that was  collected from YouTube videos. Table \ref{table:TrackingNet} shows that DomainSiam outperforms MDNet which is the second best tracker on the TrackingNet dataset. with $2\%$ in terms of precision and about $3\%$ in terms of success. DomainSiam outperforms all other trackers on TrackingNet dataset.  \\
\textbf{Results on LaSOT Dataset}\\
The average sequence length in this dataset is about 2500 frames. Table \ref{table:lasot}  shows that DomainSiam achieves the best success score with over $2\%$  from the second best tracker (DaSiam). Our tracker significantly outperforms DaSiam with $4\%$ in terms of normalized precision.\\ 
\textbf{Results on GOT10k Dataset}\\
This dataset has 180 test sequences. We tested the proposed tracker against 7 trackers as shown in Table \ref{table:GOT10k}. DomainSiam outperforms CFNet which is the best tracker in terms of Average Overlap (AO) with $4\%$. It is worth mentioning that DomainSiam achieves the best performance among all trackers in terms of success Rate (SR) at thresholds of $0.50$ and $0.75$. 

\section{Conclusions and Future Work}
\label{Conclusions }

In this paper, we introduced DomainSiam tracker, a Siamese with a ridge regression network to fully utilize the semantic and objectness  information for visual object tracking while also producing a class-agnostic. We developed a differentiable weighted-dynamic loss function to solve the ridge regression problem. The developed loss function improves the feature learning as it automatically adjusts the robustness during the training phase. Furthermore, it utilizes the activated channels which correspond to the object category label. The proposed architecture decreases the sparsity problem in Siamese networks and provides an efficient Domain-Aware feature space that is robust to appearance changes. DomainSiam does not need to be re-trained from scratch as the ridge regression network with the proposed loss function is trained separately from the Siamese network. DomainSiam with the proposed loss function  exhibits a superior convergence speed compared to other loss functions. The ridge regression network with the proposed loss function can be extended to other tasks such as object  detection and semantic segmentation.

\bibliographystyle{splncs04}

\bibliography{References}

\end{document}